\documentclass{article}
\usepackage{spconf,amsmath,graphicx}
\usepackage{hyperref}
\usepackage{cleveref}
\usepackage{amsfonts}
\usepackage{xcolor}
\newcommand{\etal}{\textit{et al}. }
\newcommand{\ie}{\textit{i}.\textit{e}. }

\title{MRI reconstruction via Cascaded Channel-wise Attention Network}
\name{Qiaoying Huang, Dong Yang, Pengxiang Wu, Hui Qu, Jingru Yi, Dimitris Metaxas}
\address{Department of Computer Science, Rutgers University, NJ, USA}
\begin{document}
%
\maketitle
\begin{abstract}
We consider an MRI reconstruction problem with input of k-space data at a very low undersampled rate.
This can practically benefit patient due to reduced time of MRI scan, but it is also challenging since quality of reconstruction may be compromised.
Currently, deep learning based methods dominate MRI reconstruction over traditional approaches such as Compressed Sensing, but they rarely show satisfactory performance in the case of low undersampled k-space data.
One explanation is that these methods treat channel-wise features equally, which results in degraded representation ability of the neural network.
To solve this problem, we propose a new model called MRI Cascaded Channel-wise Attention Network (MICCAN), highlighted by three components: (i) a variant of U-net with Channel-wise Attention (UCA) module, (ii) a long skip connection and (iii) a combined loss.
Our model is able to attend to salient information by filtering irrelevant features and also concentrate on high-frequency information by enforcing low-frequency information bypassed to the final output.
We conduct both quantitative evaluation and qualitative analysis of our method on a cardiac dataset. 
The experiment shows that our method achieves very promising results in terms of three common metrics on the MRI reconstruction with low undersampled k-space data.
Code is public available\footnote{https://github.com/charwing10/isbi2019miccan}.
\end{abstract}
\section{Introduction}
\label{sec:intro}
Magnetic resonance imaging (MRI) is widely used due to its high resolution and low radiation, but fully-sampled MRI scan requires lots of time for patients.
Therefore, MRI data is usually undersampled in the Fourier domain to improve the scan efficiency.
According to Compressed Sensing (CS) theory that assumes signal is sparse in a transform domain, reconstruction is possible even if signal is undersampling below the Nyquist rate \cite{donoho2006compressed, candes2006compressive}.
Naturally, undersampling in the transform domain introduces artifacts and significant coefficients are required to infer a de-noised image.
Such interference can be formalized as a nonlinear optimization problem, recovering the sparse coefficients and the image itself effectively. 

Lustig \etal proposed to minimize the $\ell_1$ norm of a transformed image, subject to data consistent constraints \cite{lustig2007sparse}.
Another variant is to replace $\ell_1$ with total variation (TV) penalty \cite{lustig2008compressed}.
RecPF \cite{yang2010fast} employed the alternating direction method to minimize a linear combination of $\ell_1$ and TV norm regularization.
FCSA \cite{huang2011efficient} further improved the quality of reconstructed images by weighting solutions from $\ell_1$ and TV norm subproblems in an iterative framework.
These works concentrate on designing new objective functions and optimization algorithms, but they omit the fact that reconstruction procedure is computationally inefficient and requires thousands of iterations until convergence.
Moreover, objective function largely depends on prior information of data, which is usually unknown in practical problem.

Recently, deep learning based MRI reconstruction methods are widely investigated to overcome such drawbacks due to its strong ability of feature representation.
Sun \etal \cite{sun2016deep} first proposed to use neural network in MRI reconstruction problem, where the network was derived from the iterative procedures in Alternating Direction Method of Multipliers (ADMM) algorithm \cite{boyd2011distributed} to optimize a general CS-based MRI model.
Schlemper \etal proposed a deep cascaded convolution neural network with a data consistency unit (DC-CNN) which improves the performance of MRI reconstruction \cite{schlemper2017deep}.
Recent work proposed a deep cascading of the U-net structure, which further decreases the reconstruction error \cite{sun2018joint}.

One common limitation of existing deep learning methods is that they barely show promising performance on k-space data at a very low undersampled rate.
One reason is that these methods treat channel-wise features equally, which degrades representation ability of neural network.
To resolve the problem, we propose a new model called MRI Cascaded Channel-wise Attention Network (MICCAN). 
It is featured by three components: (i) a variant of U-net with Channel-wise Attention (UCA) module, (ii) a long skip connection and (iii) a combined loss. 
The UCA module exploits channel-wise attention mechanism \cite{hu2017squeeze, zhang2018image, abhijit2018cocurrent} and aims to attend to important features related to the final goal by filtering irrelevant and noisy features.
The long skip connection allows abundant low-frequency information bypassed to the final output which enforces the model to focus on learning high-frequency information.
By adaptively reweighing features according to their inter-dependencies among feature channels, representation ability of deep network is further improved.
To the best of our knowledge, this is the first work to employ the channel-wise attention mechanism in MRI reconstruction problem.

\begin{figure*}[!t]
    \centering
    \includegraphics[width=0.95\textwidth]{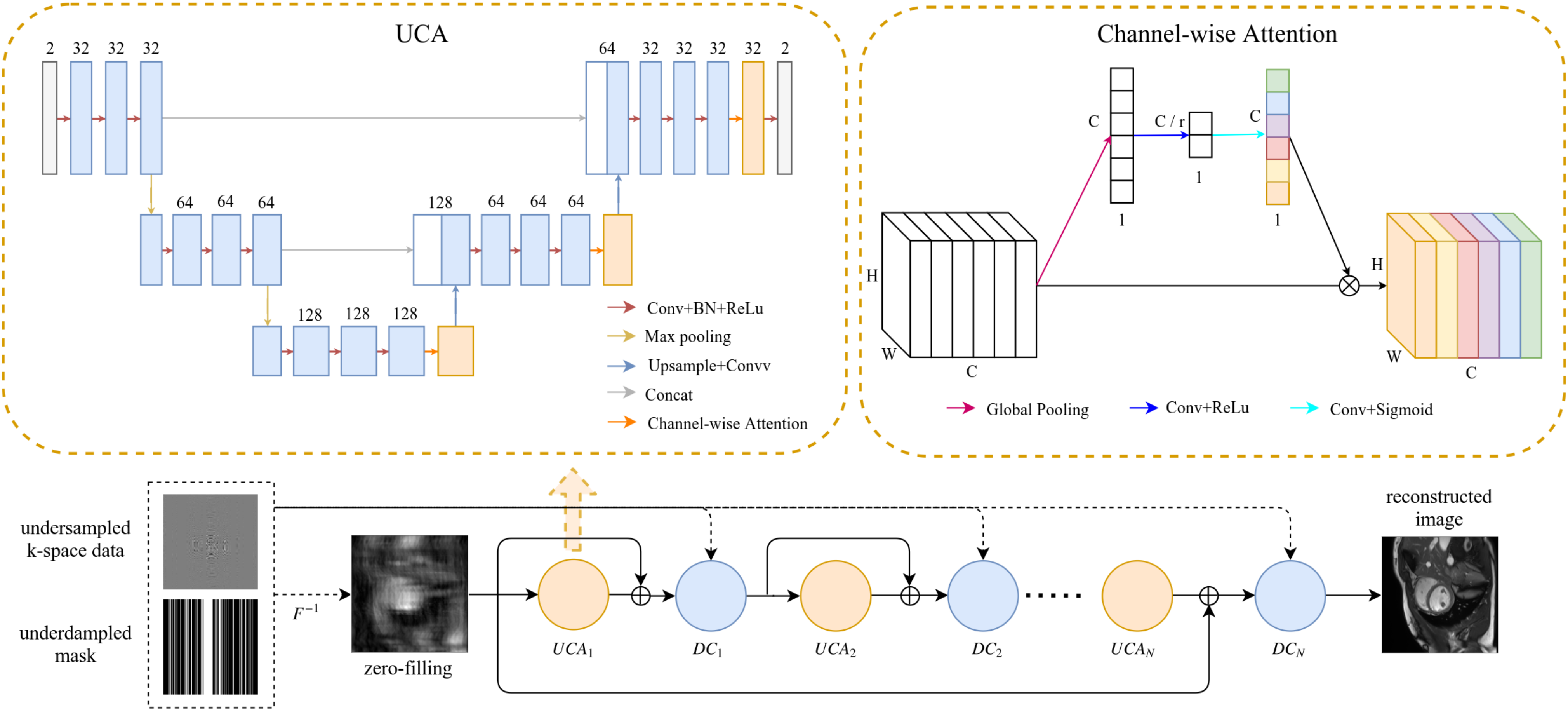}
    \caption{The framework of the proposed MICCAN network. 
The end-to-end network consists of UCA blocks, Data Consistency blocks and Channel-wise Attention units.
It takes as input the undersampled k-space data and the undersampled mask (left-most) and outputs the reconstructed MRI (right-most image).
The zero-filled reconstruction (second-left image)  works as the start point for reconstruction.
The Data Consistency unit employs the original k-space data for further refinement.}
\vspace{-0.15in}
    \label{fig:network}
\end{figure*}

\section{Methodology}
\label{sec:copyright}

In CS-based MRI reconstruction problem, the goal is to find a reconstructed image $\tilde{x}$ such that it minimizes the reconstruction error between original k-space data $y$ and Fourier transform of $x$:
\begin{equation}\label{eq:obj}
\tilde{x}=\arg\min_{x}\|F_u(x)-y\|_2^2+R(x).
\end{equation}
Here $F_u$ is an operator that transforms $x$ into Fourier domain with undersampling.
$R(\cdot)$ is a regularization term that depends on the prior information of data and general choices are $\ell_1$ or $\ell_0$.

\subsection{Preliminary}
Traditionally, the objective function of Equation \ref{eq:obj} is solved in an iterative manner that requires thousands of iteration steps until convergence.
A more efficient way is to approximate the optimal reconstructed image from undersampled k-space data via deep neural network.
One issue is that as the network goes deeper, the original information may be degraded.
Schlemper \etal \cite{schlemper2017deep} proposed a Data Consistency (DC) layer to avoid loss of information.
Basically, DC layer takes as input the reconstructed image $x_n$ from the $n$-th reconstruction block and outputs an updated ``reconstructed'' image $x_n^{dc}$.
Formally, DC layer is defined as
\begin{equation}
x_n^{dc}=\text{DC}(x_n)=F^H(\tau(F(x_n))).
\end{equation}
Here $F(x)$ is Fourier transform that takes image $x$ as input and outputs $\hat{x}$ in Fourier domain and $F^H(\hat{x})$ is inverse Fourier transform.
$\tau(\hat{x})$ is the data fidelity operation whose output has the same dimension as $\hat{x}$:
\begin{equation}\label{eq:dc1}
\tau(\hat{x})[i,j]=
\begin{cases}
\hat{x}[i,j], & (i, j)\not\in\Omega \\
\frac{\hat{x}[i,j]+v y[i,j]}{1+v}, & (i,j)\in\Omega
\end{cases}
\end{equation}
where $[i,j]$ is the matrix indexing operation, $\Omega$ is the set of sampled positions of k-space data and $v\in[0,\infty)$ is the noise level.
In the noiseless case ($v\rightarrow\infty$), we have $\hat{x}[i,j]=y[i,j]$ if $(i,j)\in\Omega$, \ie filling $\hat{x}$ with the original values of k-space data $y$ at position $(i,j)$.

\subsection{MRI Cascaded Channel-wise Attention Network}
In this section, we propose a new network called \emph{MRI Cascaded Channel-wise Attention Network} (MICCAN). 
As shown in Figure \ref{fig:network}, MICCAN mainly consists of two parts: \emph{U-net with Channel-wise Attention} (UCA) module and DC layer.
These components are cascadedly coupled together and repeat for $N$ times.
Formally, denote the $n$-th UCA module and the $n$-th DC layer respectively by UCA$_n$ and DC$_n$.
The starting point of MICCAN is undersampled k-space data $y$, which is later converted into a zero-filling image $x_0=F^H(y)$ through inverse Fourier transform $F^H$ and fed to a UCA module.
Our cascaded model can be simply formalized as
\begin{equation}
\begin{cases}
x_n=\text{UCA}_n(x_{n-1}^{dc})+x_{n-1}^{dc} \\
x_n^{dc}=\text{DC}_n(x_n) \\
\end{cases}
(n=1,\ldots,N)
\end{equation}
where $x_0^{dc}$ is initialized as $x_0$.
The final reconstructed image of MICCAN, namely $x_N^{dc}$, is produced by the last DC layer (DC$_N$).


\subsubsection{U-net with Channel-wise Attention}\label{sec:attention}
In the previous work on reconstruction problem, deep learning based methods have two major issues.
First, they treat each channel-wise feature equally, but contributions to the reconstruction task vary from different feature maps. 
Second, receptive field in convolutional layer may cause to lose contextual information from original images, especially high-frequency components that contain valuable detailed information such as edges and texture.
Therefore, inspired by \cite{abhijit2018cocurrent}, we develop the UCA module by introducing an attention mechanism that filters the useless features and enhance the informative ones.
The attention technique is only applied in the decoder part. 
The intuition is that features of the decoder are extracted from coarse to fine feature-maps of multiple scales via skip connection.
The attention module filters salient and prunes irrelevant and noisy features such that allows model parameters in shallower layers to be updated mostly that are relevant to a given task. 
To the best of our knowledge, this is the first work to employ channel-wise attention to MRI reconstruction problem.

Specifically, we use global average pooling to extract the channel-wise global spatial information to vector $z\in \mathbb{R}^{C}$, whose $c$ dimension is defined as
\begin{equation}
z_c = \frac{1}{H\times W}\sum_{i=1}^H\sum_{j=1}^{W}f_c[i,j] ~~(c=1,\ldots,C)
\end{equation}
where $f_c\in \mathbb{R}^{W\times H}$ is the feature map in the $c$-th channel.
Such operation squeezes the spatial information of the whole image into a vector length of $C$.
To further extract feature related to the final task, we introduce another gating mechanism as follows:
\begin{equation}
\hat{x}_c = \sigma(\delta(z*\mathbf{W}_1)*\mathbf{W}_2)\odot f_c,
\end{equation}
where ``$*$'' is convolutioj operator and $\delta(\cdot)$ is ReLU activation function to encode the channel-wise dependencies. 
$\mathbf{W}_1$ is a kernel in the first convolutional layer that reduces the $C$-dimentional feature vector into $C/r$.
On the contrary, kernel $\mathbf{W}_2$ increases feature size back to $C$. 
Sigmoid function $\sigma(\cdot)$ is used to compute weighted attention map, which is later applied to rescaling the input feature $f_c$.
Based on this attention mechanism in the UCA module, our model MICCAN achieves very promising results and outperforms several state-of-the-art methods. 

\subsubsection{Learn Global Residual by Long Skip Connection}
To address the problem of vanishing low frequency in deep learning based MRI reconstruction, we utilize a long skip connection from the zero-filling image to the final reconstruction block. 
Specifically, we replace the residual in the last UCA module, namely $x_N=\text{UCA}_N(x_{N-1}^{dc})+x_{N-1}^{dc}$, with a long skip connection:
\begin{equation}
x_N=\text{UCA}_N(x_{N-1}^{dc})+x_0
\end{equation}
This simple modification is used to learn the global residual and to stabilize the gradient flow in deep residual network.

\subsubsection{Combined Loss Function}\label{sec:loss}
Common choice of loss function for reconstruction problem is $\ell_2$, but the resulting reconstructed image is of low quality and lacks high frequency detail.
Therefore, we propose to use a combination of loss functions including $\ell_1$ loss and perceptual loss $\ell_{p}$\cite{johnson2016perceptual}. 
Given target image $x_s$ and reconstructed image $x\doteq x_N^{dc}$ of MICCAN parameterized by $\theta$, the combined loss is defined as
\begin{equation}
\ell^{\theta}(x,x_s)=\lambda_1\ell_{1}^{\theta}(x, x_s)+ \lambda_{p}\ell_p^{\theta}(x, x_s)
\end{equation}
where
\begin{equation}
\ell_{1}^{\theta}(x, x_s)=\|x-x_s\|_1
\end{equation}
\begin{equation}
\ell_{p}^{\theta}(x, x_s)=\sum_{k=1}^K \|\phi_{VGG}^k(x)-\phi_{VGG}^k(x_s)\|^2_2
\end{equation}
where $\lambda_1$ and $\lambda_{p}$ are weighing factors for two losses, $\phi_{VGG}^k(\cdot)$ represents features of the $k$-th activation layer in VGG network.
Note that perceptual loss $\ell_{p}$ minimizes the $\ell_2$ distance between reconstruction image and target image in $K$ different feature spaces, or equivalently it encourages the predicted image to be perceptually similar to the target image.

\begin{figure*}
    \centering
    \includegraphics[width=0.95\textwidth]{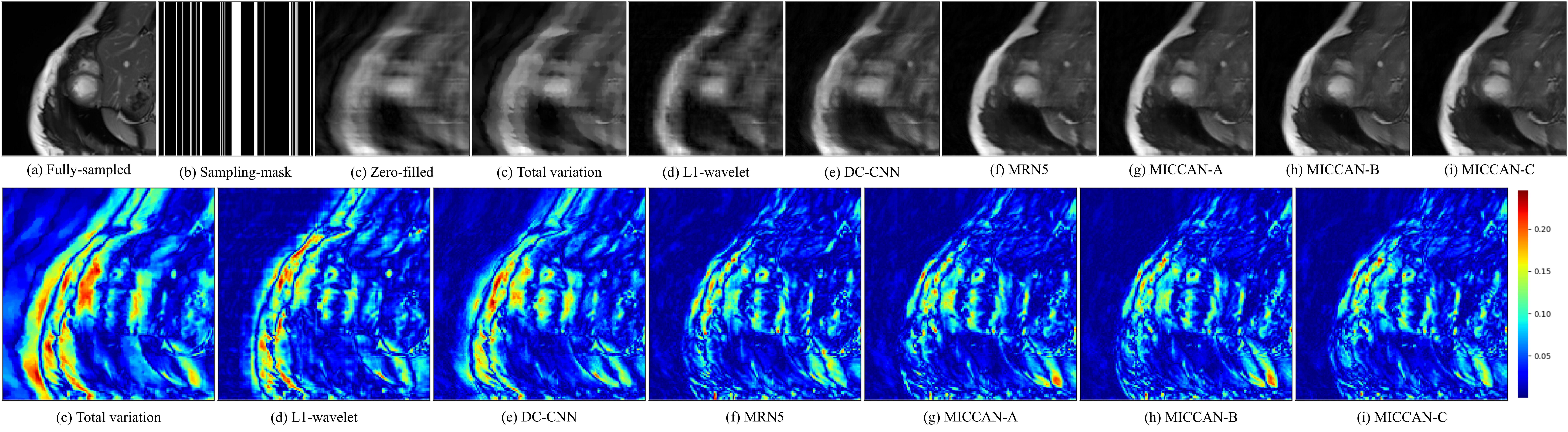}
    \caption{Visualization of reconstructed images and reconstruction errors with k-space data at undersampled rate $12.5\%$ }
    \label{fig:recresult}
    \vspace{-0.15in}
\end{figure*}

\section{Experiments}
\label{sec:exp}
To evaluate the effectiveness of our proposed model, we compare it with several state-of-the-art approaches on a simulated cardiac k-space dataset.

\vspace{0.5em}\noindent\textbf{Methods}\hspace{1em}
We evaluate following methods in the experiment.
Two traditional CS-MRI methods include $\ell_1$-wavelet \cite{lustig2007sparse} and TV norm \cite{lustig2008compressed}, which are implemented in the BART toolbox \cite{tamir2016generalized}.
Two deep learning based models include DC-CNN\cite{schlemper2017deep} and MRN5\cite{sun2018joint}.
The only difference between our proposed MICCAN and MRN5 is that MRN5 does not used attention module and long skip connection.
We also consider three variants of our methods: MICCAN with $\ell_2$ loss (MICCAN-A), MICCAN with combined loss proposed in Section \ref{sec:loss} (MICCAN-B) and MICCAN with both long skip connection and combined loss (MICCAN-C).
The same as \cite{hu2017squeeze, abhijit2018cocurrent}, we set the reduction ratio $r$ that is mentioned in Section \ref{sec:attention} as $8$ for all our MICCAN models.
For the combined loss, we set $\lambda_1$ as $10$ and $\lambda_p$ as $0.5$.
As the shallow features encode details of images, we compute the perceptual loss of layer $relu1$-$2$, $relu2$-$1$, $relu2$-$2$ and $relu3$-$1$ of the VGG-16 network.
All deep learning models are implemented using PyTorch and trained on NVIDIA K80.
Learning rate is initialized as $10^{-4}$ with decreasing rate of $0.5$ for every 15 epochs. 
The training batch is $8$ and the maximum number of epochs is $50$.
For fair comparison, we follow \cite{schlemper2017deep, sun2018joint} by setting the number of reconstruction blocks $N$ as $5$. 

\vspace{0.5em}\noindent\textbf{Dataset}\hspace{1em}
A cardiac MRI dataset with $15$ subjects is adopted in our experiments.
We randomly choose $10$ subjects as training set, $2$ subjects as validation set and the rest $3$ subjects as test set.
We follow the k-space data simulation method proposed in \cite{jung2007improved} .
It assumes the sampled mask follows a zero-mean Gaussian distribution and the Cartesian undersampling method is adopted, also keeps the eight lowest spatial frequencies.
Our model is evaluated on data with undersampled rate at $12.5\%(\text{acceleration rate } 8 \times)$.

\begin{table}[!t]
\caption{Reconstruction results of the proposed MICCAN models and other methods.}
\centering
\small
\begin{tabular}{p{1.6cm}|p{1.7cm}p{1.9cm}p{1.7cm}}
\hline
Methods & NRMSE & PSNR & SSIM \\
\hline
\hline
TV  & $0.1087_{\pm 0.0346}$ & $19.7388_{\pm 2.8942}$ & $0.5923_{\pm 0.0898}$ \\
$\ell_1$-wavelet  & $0.0753_{\pm 0.0185}$ & $22.7054_{\pm 2.0255}$ & $0.6333_{\pm 0.0553}$ \\
DC-CNN           & $0.0587_{\pm 0.0118}$ & $24.7993_{\pm 1.7612}$ & $0.6612_{\pm 0.0582}$ \\
MRN5             & $0.0427_{\pm 0.0076}$ & $27.5373_{\pm 1.5971}$ & $0.7851_{\pm 0.0297}$ \\
\hline
MICCAN-A         & $0.0402_{\pm 0.0072}$ & $28.0664_{\pm 1.6231}$ & $0.8005_{\pm 0.0301}$ \\
MICCAN-B         & $0.0391_{\pm 0.0074}$ & $28.3283_{\pm 1.7080}$ & $\mathbf{0.8214_{\pm 0.0327}}$ \\
MICCAN-C         & $\mathbf{0.0385_{\pm 0.0073}}$ & $\mathbf{28.4489_{\pm 1.6953}}$ & $0.8198_{\pm 0.0302}$ \\
\hline
\end{tabular}
\label{tab:exp}
\end{table}

\subsection{Quantitative Evaluation}
In this experiment, we quantitatively evaluate all models with three widely used measurements: normalized root square mean error (NRMSE), peak signal-to-noise ration (PSNR) and the structural similarity index measure (SSIM). 
The results are shown in Table \ref{tab:exp} and we basically observe following three trends.
First, all our MICCAN variants outperform other baseline models. 
This mainly attributes to the attention mechanism proposed in Section \ref{sec:attention}.
Even trained with $\ell_2$ loss, MICCAN-A achieves NRMSE of $0.0402$, PSNR of $28.0664$ and SSIM of $0.8005$ that beats MRN5 model.
Second, MICCAN-B model that is trained with combined loss gains better result than MICCAN-A.
For example, MICCAN-B has $2.7\%$ decrease in NRMSE compared with MICCAN-A and also achieves the best SSIM value of $0.8214$.
This indicates that our combined loss is better than the $\ell_2$ loss, making image less blur and keeping the perceptual details.
Third, with long skip connection, MICCAN-C further improves reconstruction performance with the lowest NRMSE of $0.0385$ and the highest PSNR value of $28.4489$.
Overall, all these results demonstrate the effectiveness of the channel-wise attention modules and the proposed combined loss.

\subsection{Qualitative Analysis}
In this experiment, we visualize the reconstructed image and reconstruction errors on a test sample and qualitatively analyze the results of all methods.
As shown in Figure \ref{fig:recresult}, most of the baseline methods cannot completely recover detail of the image and suffer from severe blurring artifacts. 
In contrast, our three MICCAN methods eliminate most blurring artifacts and recover more details from low undersampled k-space data. 
Furthermore, for the transitional methods such as $\ell_1$-wavelet and TV, the reconstructed images are similar to the zero-filling image and suffer from heavy blurring artifacts.
We highlight that our MICCAN achieves best results and with much fewer reconstruction errors.

\section{Conclusion}
In this paper, we propose a robust and effective model called MICCAN for MRI reconstruction from very low undersampled k-space data.
We devise a variant of U-net with channel-wise attention  as reconstruction block together with long skip connection technique and a combined loss function. 
We evaluate our method and current state-of-the-art approaches on a cardiac benchmark.
The experimental results show that our method outperforms all other compared methods and achieves very promising results under the condition of low undersampled rate.

\label{sec:con}

\bibliographystyle{IEEEbib}
\bibliography{refs}
\end{document}